\documentclass[11pt, a4paper, logo, copyright, nonumbering]{xiaomi}

\usepackage[authoryear, sort&compress, round]{natbib}
\usepackage{dblfloatfix}
\usepackage{ulem}
\usepackage{caption}
\usepackage{dramatist}
\usepackage{xspace}
\usepackage{pifont} 
\usepackage{multirow}
\usepackage{tcolorbox}
\usepackage{xltabular}
\usepackage{longtable}
\usepackage{hyperref}
\usepackage{cleveref}
\usepackage{wrapfig}
\usepackage{graphicx}

\usepackage{amsfonts}
\usepackage{amsmath}
\usepackage{amssymb}
\usepackage{booktabs}
\usepackage{lineno}
\usepackage{multirow}
\usepackage{adjustbox}

\usepackage[bottom]{footmisc}

\usepackage{CJKutf8}
\usepackage{setspace}
\usepackage{makecell}
\usepackage{graphicx}
\usepackage{subcaption}
\usepackage{multicol} 

\definecolor{xiaomiorange}{HTML}{FF6901}

\allowdisplaybreaks[4]

\title{\centering Reinforcement Learning for Chain of Thought Compression with One-Domain-to-All Generalization}

\author{
	{\normalfont\sffamily\fontsize{11}{15}\selectfont Hanyu Li$^{\dagger\S}$ ~~~~~Jiangshan Duo$^{\dagger\ddagger}$ ~~~~~Bofei Gao$^{\dagger\ddagger}$ ~~~~~Liang Zhao$^{\dagger}$} \\
	{\normalfont\sffamily\fontsize{11}{15}\selectfont Hailin Zhang$^{\dagger}$ ~~~~~Sujian Li$^{\ddagger}$ ~~~~~Xiaotie Deng$^{\S}$ ~~~~~Fuli Luo$^{\dagger}$} \\
	\vskip10pt
	{\normalfont\sffamily\fontsize{11}{15}\selectfont $^{\S}$CFCS, School of Computer Science, Peking University} \\
	{\normalfont\sffamily\fontsize{11}{15}\selectfont $^{\ddagger}$State Key Laboratory of Multimedia Information Processing, School of } \\
	{\normalfont\sffamily\fontsize{11}{15}\selectfont  Computer Science, Peking University} \\
	{\normalfont\sffamily\fontsize{11}{15}\selectfont $^{\dagger}$LLM-Core Xiaomi}
	\vskip10pt
}

\begin{abstract}
Chain-of-thought reasoning in large language models can trigger an ``overthinking trap'': longer rollouts raise cost and latency yet often yield unreliable accuracy gains. Existing methods use global, static controls that may suppress needed reasoning. We propose mastery-gated, sample-level, soft reinforcement learning compression that penalizes long rollouts only when the model already solves the problem and has produced a shorter rollout. Across benchmarks, it cuts response length by 20-40\% with comparable or higher accuracy and generalizes across domains: a model trained on math spontaneously shortens unseen tasks (code, instruction following, general-knowledge QA) without hurting accuracy. We further show two-way transfer between non-agent CoT and tool-use agents: non-agent training reduces SWE-Bench Verified rounds by 13\%, while compressing a thinking agent cuts SWE trajectories by 67\% tokens and 52\% rounds and shortens non-agent outputs by up to 44\%. Compression is thus not cosmetic brevity, but an inherent computation policy --- what to keep, and what to forget.

\end{abstract}

\begin{document}

{
    \bgroup
    \setlength{\parindent}{0pt}
    \vspace*{3pt} 
    \begin{adjustwidth}{0pt}{0pt}  
    \begin{center} 
    {\titlefont Reinforcement Learning for Chain of Thought Compression with One-Domain-to-All Generalization \par}
    \vskip5pt
    {
	{\normalfont\sffamily\fontsize{11}{15}\selectfont Hanyu Li$^{\dagger\S\star}$ ~~~~~Jiangshan Duo$^{\dagger\ddagger}$ ~~~~~Bofei Gao$^{\dagger\ddagger}$ } ~~~~~
	{\normalfont\sffamily\fontsize{11}{15}\selectfont Hailin Zhang$^{\dagger}$ \\ Sujian Li$^{\ddagger}$ ~~~~~Xiaotie Deng$^{\S}$ ~~~~~Liang Zhao$^{\dagger\diamond}$} \\
	\vskip10pt
	{\normalfont\sffamily\fontsize{11}{15}\selectfont $^{\S}$CFCS, School of Computer Science, Peking University} \\
	{\normalfont\sffamily\fontsize{11}{15}\selectfont $^{\ddagger}$State Key Laboratory of Multimedia Information Processing, School of } \\
	{\normalfont\sffamily\fontsize{11}{15}\selectfont  Computer Science, Peking University} \\
	{\normalfont\sffamily\fontsize{11}{15}\selectfont $^{\dagger}$LLM-Core Xiaomi}
	\vskip10pt
    }
    \end{center}
    \end{adjustwidth}
    \egroup
    {\abscontent}
    \thispagestyle{firststyle} 
}

\begin{figure}[ht]
    \centering
    \includegraphics[width=.9\textwidth]{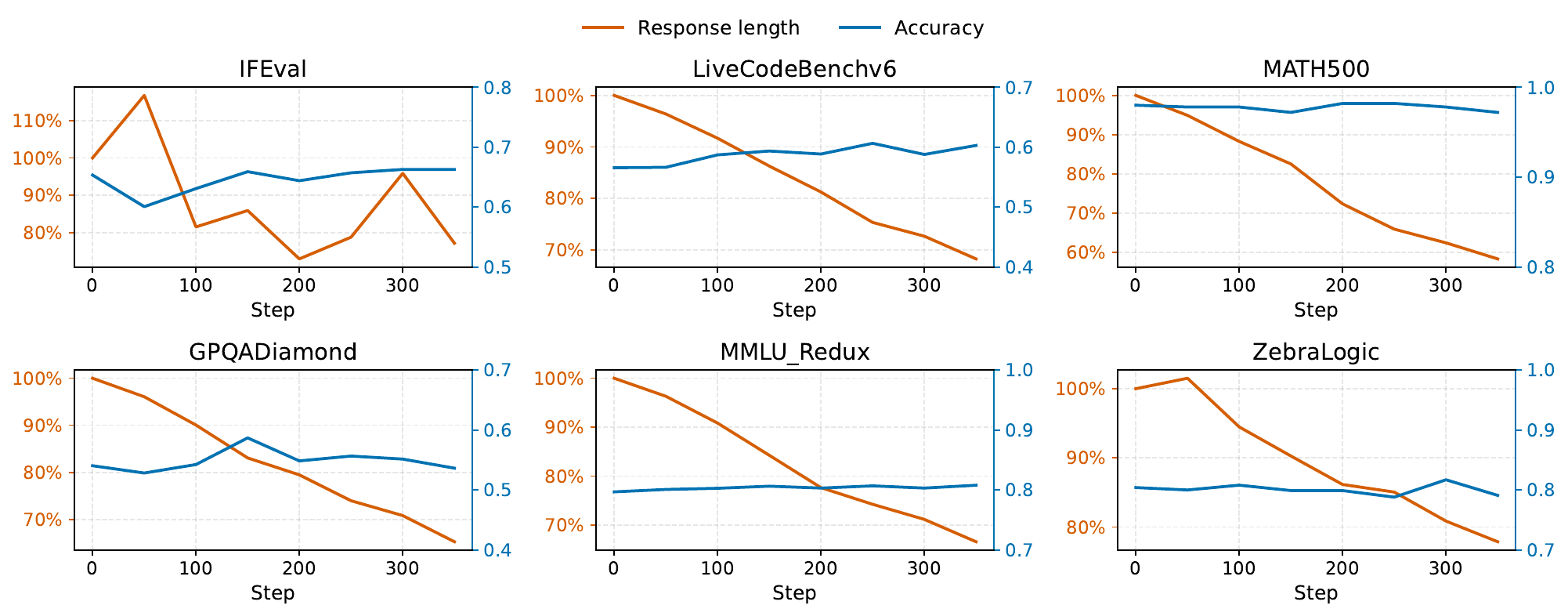}
    \caption{
    Overview of our compression method across diverse tasks.
    Each panel shows one benchmark: the orange curve is response length (tokens, normalized to 100\% at that benchmark's starting point) and the blue curve is the core metric (accuracy or score), illustrating that compression training \textit{only} on math problems broadly shortens chains of thought while maintaining or improving downstream performance.
    }
    \label{fig:benchmark-length-accuracy-grid}
\end{figure}

\renewcommand{\thefootnote}{\fnsymbol{footnote}}
\footnotetext[0]{$^{\star}$Contribution during internship at Xiaomi LLM-Core Team.}
\footnotetext[0]{$^{\diamond}$Corresponding author.}
\renewcommand{\thefootnote}{\arabic{footnote}}

\section{Introduction}

Generating long chains of thought (CoT) has significantly improved the complex reasoning abilities of large language models~\citep{weiChainofthoughtPromptingElicits2022,guoDeepSeekR1IncentivizesReasoning2025}. However, this success introduces an ``overthinking trap,'' characterized by high computational cost, increased latency, and verbose reasoning paths that do not always guarantee higher accuracy~\citep{teamKimiK15Scaling2025,yuDAPOOpenSourceLLM2025}. We posit that true intelligence manifests as compression. Once a model masters a problem-solving path, a more efficient representation must exist. We argue that compression is not merely a trade-off for efficiency but an indicator of internalized, mature capability.

Prevailing approaches, including reinforcement learning with various length penalties~\citep{teamKimiK15Scaling2025,yuDAPOOpenSourceLLM2025} and various forms of distillation~\citep{hsiehDistillingStepbyStepOutperforming2023,daiCaptureKeyReasoning2025,liTeachingSmallLanguage2024}, have laid a crucial foundation. They demonstrate the feasibility of compressing CoT and offer valuable mechanisms for controlling computational budgets, often achieving effective accuracy-efficiency trade-offs. However, these methods typically rely on \textit{global} and \textit{static} controls, such as a uniform penalty or a fixed truncation limit. 

Such ``one-size-fits-all'' policies risk operating at the wrong \textit{unit}: they do not adapt to the heterogeneous difficulty of individual samples and may inadvertently penalize necessary, extended reasoning. Furthermore, they often apply compression at the wrong \textit{time}, forcing shorter solutions before the model has reliably mastered the problem and causing the reward hacking problem. We contend that effective length control must be sample-level, gated, and soft, rather than global, ungated, and hard.

We propose a minimal intervention: apply soft compression \textit{only} to samples the model has already mastered, and \textit{only} on reasoning paths that exceed a ```safe length'' derived from its own correct solutions. This approach ensures that difficulty, which we distinguish from length, is respected. A model must first learn to solve a problem reliably; only then is it encouraged to find a shorter path. We penalize only the excessive verbosity on problems the model demonstrably understands, rather than forcing truncation on problems it is still reasoning.

Our experiments reveal several key findings. First, this sample-level, gated, soft compression framework significantly outperforms global penalties, while hard truncation methods fail catastrophically. The gating mechanism --- compressing only after mastery is achieved --- proves essential for maintaining robust performance. Second, this compression skill generalizes: models trained to compress reasoning on a specific domain (e.g., mathematics) spontaneously reduce verbosity across all tasks. Third, we identify a stable, multi-stage training recipe which can reduce average inference length by 20--40\% with comparable or even improved accuracy, a robust result across model structures, scales, and tasks. Finally, we show that this learned notion of ``efficient computation'' extends beyond single-turn CoT: it transfers to tool-use agents and even exhibits two-way generalization between non-agent reasoning and agent trajectories.

Based on these findings, we argue that reasoning compression should be elevated from an optimization trick to a standard, distinct phase of post-training. We propose a stable curriculum: first, enhance accuracy; second, apply safe, gated compression with early stopping; finally, conduct another accuracy-enhancing phase to recover or further boost performance. This transforms the principle of ``compression as intelligence'' into a repeatable engineering process, capable of significantly improving inference efficiency by learning \textit{when} and \textit{how} to compress.

Our contributions are threefold. We (1) propose a minimal, stable, and sample-level gated compression paradigm, formalized as a practical post-training recipe; (2) systematically validate the mechanisms, from motivation to empirical phenomena, providing reproducible evidence of its dynamics, generalization, and failure modes, leading to a robust early stopping strategy; (3) reframe compression from a mere cost-optimization technique to a measurable process of capability internalization, and show it as an agent-agnostic efficiency skill that applies to both pure reasoning and tool-use settings.

\section{Related Work}\label{sec:related-work}

\paragraph{Reinforcement learning for length control.} Most prior work relies on global, static policies-hard truncation (ThinkPrune~\citep{houThinkPrunePruningLong2025}), truncation-penalized RL (DLER~\citep{liuDLERDoingLength2025}), length-harmonizing fine-tuning (O1-Pruner~\citep{luoO1PrunerLengthHarmonizingFineTuning2025}), constraint-following RL (L1~\citep{aggarwalL1ControllingHow2025}), sample-level non-gated length penalties RL in Kimi K1.5~\citep{teamKimiK15Scaling2025}, and global non-gated length penalties in DAPO~\citep{yuDAPOOpenSourceLLM2025} --- whereas we apply sample-level, gated, soft compression that triggers only after mastery and only beyond a self-derived safe length.

\paragraph{Distillation, pruning, and model merging.} CoT distillation (Distilling Step-by-Step~\citep{hsiehDistillingStepbyStepOutperforming2023}, EDIT~\citep{daiCaptureKeyReasoning2025}, D\&R Distillation~\citep{liTeachingSmallLanguage2024}), token/computation pruning and compression (Efficient Long CoT~\citep{wangEfficientLongCoT2025}, DLCoT~\citep{luoDeconstructingLongChainofThought2025}, TokenSkip~\citep{xiaTokenSkipControllableChainofThought2025}, LightThinker~\citep{zhangLightThinkerThinkingStepbyStep2025}, Prune-on-Logic~\citep{zhaoCanPruningImprove2025}), and model merging (Thinking Spectrum~\citep{lanThinkingSpectrumEmpirical2025}, model merging for length-controlled reasoning~\citep{wuUnlockingEfficientLongtoShort2025}) pursue similar accuracy-efficiency goals, yet our loop is \textit{on-policy} and compresses only when the model has demonstrably mastered the instance.

\section{Method}
\label{sec:method}

We adopt the standard reinforcement learning with verifiable rewards (RLVR) training strategy~\citep{guoDeepSeekR1IncentivizesReasoning2025}. We compress only where the model already knows the answer. For each training sample, tokens generated beyond a per-sample ``safe length'' are softly penalized, but only after the model has demonstrated reliable mastery in solving that sample.

Let $x$ be an input problem, $y$ be a full generated response (which includes both the chain-of-thought reasoning and the final answer), and $l(y)$ be its total token length. We measure the full response length specifically to prevent the model from hacking the reward by moving reasoning content into an unpenalized final answer. A verifier provides a binary correctness reward $r^{\text{cor}}(x,y) \in \{0, 1\}$. All statistics used for length control are computed on-policy, meaning they are derived from the most recent rollouts sampled by the current model policy, not from a static or stale dataset.

For each sample $x$ at the current training step, we generate $N$ on-policy rollouts, $\{y_1, ..., y_N\}$. The sample's current-step \textit{passrate} is defined as the average correctness reward over these rollouts:

\[\hat{p}(x) = \frac{1}{N}\sum_{t=1}^{N} r^{\text{cor}}(x, y_t).\]
The compression objective is activated only when this gate is passed, which we define as $\hat{p}(x) = 1.0$. This mastery gate serves as a way to avoid reward hacking: it prevents the model from shortening reasoning on problems it has not yet learned, removing the incentive to provide very short responses by intentionally failing a problem.

When the gate is open (i.e., $\hat{p}(x) = 1.0$), we define two sample-specific length targets based on the subset of correct rollouts from the current step. The targets are: a safe lower bound, $L^{\text{start}}(x) = \text{median}\{l(y_t)\}$, and a penalty upper bound, $L^{\text{max}}(x) = \max\{l(y_t)\}$.

We then apply a soft penalty $r^{\text{len}}(x,y)$ as a piecewise linear function similar to DAPO~\citep{yuDAPOOpenSourceLLM2025}. Responses shorter than the safe median length ($L^{\text{start}}$) are not penalized. Responses longer than the maximum observed correct length ($L^{\text{max}}$) receive a full penalty of $-1$. Responses with lengths between these two bounds are penalized linearly:
\begin{equation}
r^{\text{len}}(x,y) = \begin{cases}
    0, & \text{if } l(y) \le L^{\text{start}}(x), \\
    -1, & \text{if } l(y) > L^{\text{max}}(x), \\
    -\frac{l(y) - L^{\text{start}}(x)}{L^{\text{max}}(x) - L^{\text{start}}(x)}, & \text{if } L^{\text{start}}(x) < l(y) \le L^{\text{max}}(x).
\end{cases}
\end{equation}
In the edge case where  $L^{\text{start}}(x) = L^{\text{max}}(x)$, we let $r^{\text{len}}(x,y)=0$, meaning no length penalty on these rollouts.

The final shaped reward $\tilde{r}(x,y)$ is the combination of the verification reward and this gated length penalty:
\begin{equation}
\tilde{r}(x,y) = r^{\text{cor}}(x,y) + I\{\hat{p}(x) = 1.0\} \cdot r^{\text{len}}(x,y).
\end{equation}
where $I\{\cdot\}$ is the indicator function. This design ensures that (i) problem with incorrect rollout ($r^{\text{cor}}=0$) never receive a length penalty, preserving the priority of correctness, and (ii) we avoid the hard truncation limits that can cause abrupt distributional shifts and optimization instability.

To balance the objectives of ``learn to solve'' and ``learn to compress,'' the dataloader implements a dynamic sampler. At each step, it partitions samples into two sets based on their current mastery state: compressible (those with $\hat{p}(x) = 1.0$) and non-compressible. Each training batch is constructed to draw a target mixture $\rho \in (0, 1)$ of samples from the compressible set and $1 - \rho$ from the rest. This keeps optimization pressure focused on shortening mastered cases while preserving capacity for capability learning on unsolved problems and avoiding catastrophic forgetting.

We optimize the model using the GRPO algorithm family (including GRPO~\citep{shaoDeepSeekMathPushingLimits2024}, DAPO~\citep{yuDAPOOpenSourceLLM2025}, Dr. GRPO~\citep{liuUnderstandingR1ZeroLikeTraining2025}, and modifications for mixture-of-experts models~\citep{zhengGroupSequencePolicy2025,maStabilizingMoEReinforcement2025}) of policy-gradient methods, which are robust for our setting. Additionally, we apply early stopping during the compression phase: we monitor validation accuracy and average response length, stopping at the ``pre-collapse'' optimum --- the point where accuracy first shows a sustained decline while length is still falling. This reliably avoids the over-compression regime.

This method forms a simple and stable post-training recipe: first, train for accuracy with only the binary verification reward (accuracy stage); second, enable the mastery gate and soft penalty to compress (compression stage), stopping early at the pre-collapse optimum; finally, resume a new accuracy stage from the compressed checkpoint to recover any minor accuracy dips and further improve performance on the new, more efficient reasoning distribution. This entire accuracy $\rightarrow$ compression loop can be run once as a standard post-training pass or repeated.

\section{Experiments}\label{sec:experiments}

\subsection{Experimental Setup}
We base our experiments on MiMo-7B-RL~\citep {xiaomiMiMoUnlockingReasoning2025}, a 7B dense thinking model with strong mathematical reasoning capabilities but relatively long CoTs. During the training, we set its context length to 64K tokens to accommodate long-form mathematical reasoning.  We use 113K math problems with ground truth answers for RLVR training, sourced from various public datasets.

Our training algorithm is DAPO~\citep{yuDAPOOpenSourceLLM2025}, a GRPO-family policy gradient method, with clipping parameters $\text{clip\_low}=0.2$ and $\text{clip\_high}=0.27$. We implement a dynamic sampling and mixing strategy for training: each batch is constructed so that 90\% of samples are used for standard capability training (pursuing correctness). The remaining 10\% of samples, drawn from those the model has already mastered ($\text{passrate}=1.0$), are used for compression training via our sample-level soft penalty. Unlike DAPO, we do not filter out samples with passrates $0$ or $1$ when constructing the training set. The batch size is set to 256 and the rollout size is $n=8$. We use AdamW optimizer~\citep{loshchilovDecoupledWeightDecay2018} for training. The learning rate is set to $3\times 10^{-6}$ with a $10$-step warmup. We use a training temperature of $1.0$ and a validation temperature of $0.6$, with top-$p$ set to $1.0$ for both.

We compare our method against three baselines:
\begin{itemize}
    \item \textbf{DAPO-lite}: A global soft penalty applying the DAPO penalty to the top 20\% longest rollouts in the training set ($L_{\text{start}}=26000$).
    \item \textbf{DAPO-heavy}: A more aggressive global penalty on the top 40\% longest rollouts ($L_{\text{start}}=18000$).
    \item \textbf{Sample-level hard truncation}: A baseline that, upon mastery, truncates responses at the target length instead of applying a soft penalty.
\end{itemize}
We evaluate performance on the AIME24~\citep{MathaiAime24Datasets} and AIME25~\citep{MathaiAime25Datasets} benchmarks, reporting correctness and average response length, with all metrics computed over 32 samples per test case and averaged (avg@32). 

\subsection{Main Results}

\begin{figure}[htb]
    \centering
    \begin{subfigure}[c]{.34\textwidth}
        \centering
        \includegraphics[width=\linewidth]{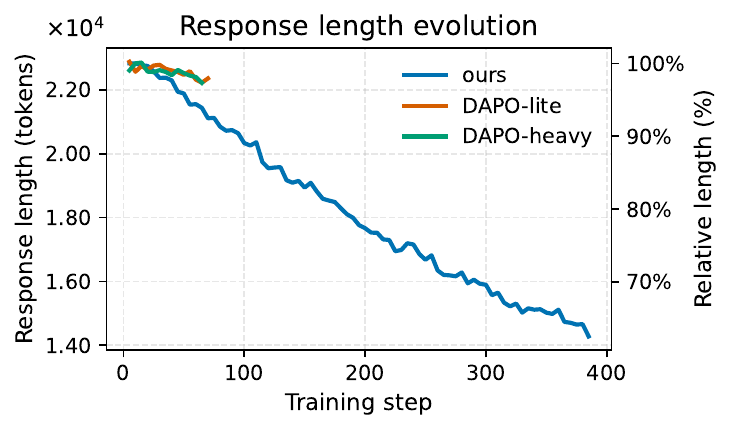}
    \end{subfigure}
    \hfill
    \begin{subfigure}[c]{.30\textwidth}
        \centering
        \includegraphics[width=\linewidth]{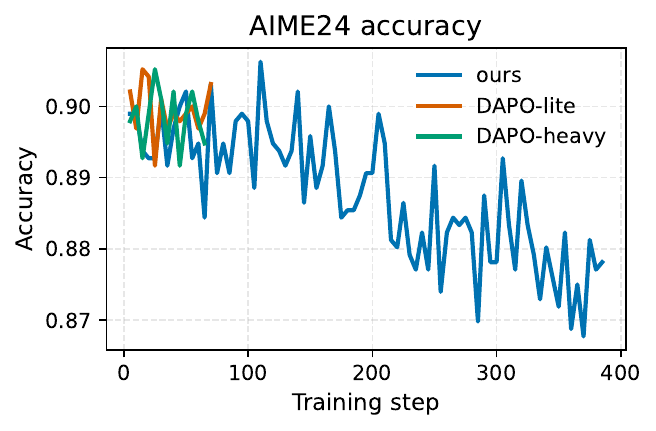}
    \end{subfigure}
    \hfill
    \begin{subfigure}[c]{.30\textwidth}
        \centering
        \includegraphics[width=\linewidth]{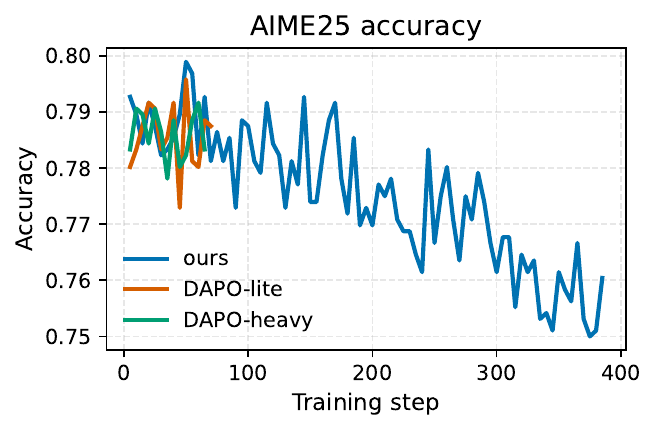}
    \end{subfigure}
    \caption{
    Training dynamics of response length (measured in tokens) and mathematical accuracy for our method (\textsc{ours}) versus global soft penalties (DAPO-lite and DAPO-heavy).
    Left: validation response length in tokens, with absolute values shown in scientific notation on the left axis and percentages relative to the initial length of \textsc{ours} on the right axis.
    Middle and right: validation accuracies on AIME24 and AIME25 as training progresses, where \textsc{ours} maintains performance while global penalties compress more slowly and offer weaker accuracy--efficiency trade-offs.
    }
    \label{fig:length-accuracy-dynamics}
\end{figure}

First, our method significantly outperforms global soft penalties. As shown in \Cref{fig:length-accuracy-dynamics}, our sample-level approach achieves rapid and steady length reduction without immediate performance degradation. On the other hand, the global soft penalty (DAPO-lite and DAPO-heavy) reduce length at a much slower rate, demonstrating the inefficiency of a ``one-size-fits-all'' policy.

\begin{figure}[htb]
    \centering
    \begin{subfigure}[t]{.38\textwidth}
        \centering
        \includegraphics[width=\linewidth]{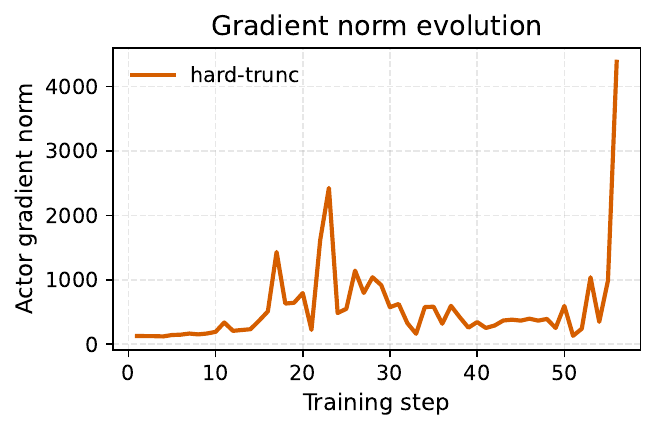}
    \end{subfigure}
    \quad
    \begin{subfigure}[t]{.38\textwidth}
        \centering
        \includegraphics[width=\linewidth]{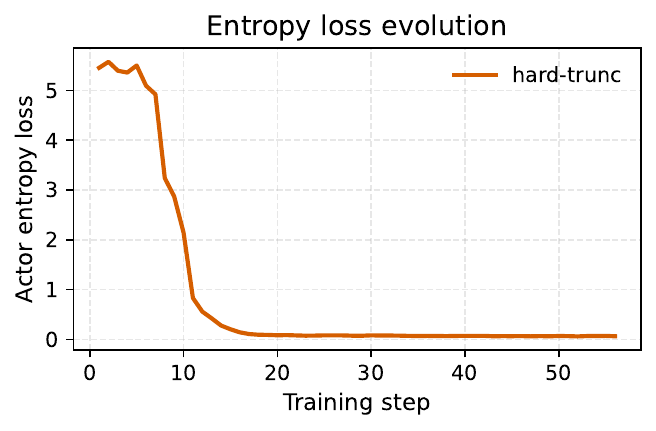}
    \end{subfigure}\\[2pt]
    \begin{subfigure}[t]{.38\textwidth}
        \centering
        \includegraphics[width=\linewidth]{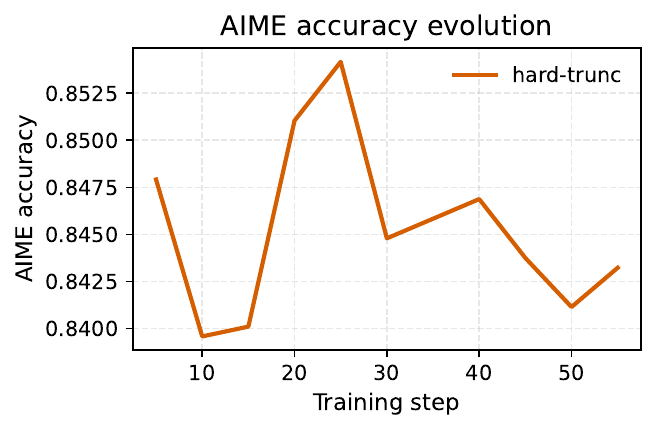}
    \end{subfigure}
    \quad
    \begin{subfigure}[t]{.38\textwidth}
        \centering
        \includegraphics[width=\linewidth]{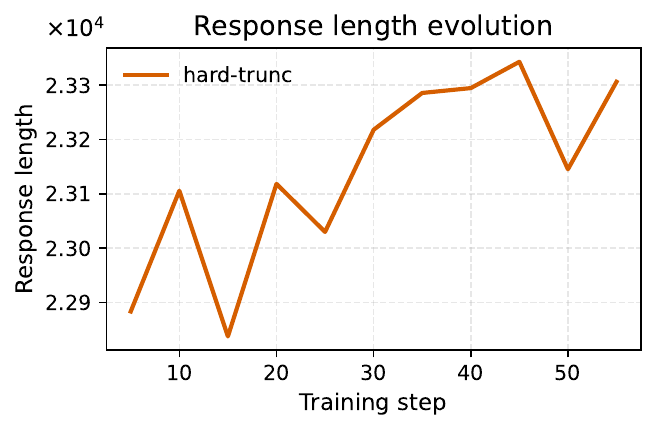}
    \end{subfigure}
    \caption{
    Training dynamics of the sample-level hard truncation baseline.
    Top: actor gradient norm and entropy loss, where truncation leads to gradient explosion and collapse of entropy, indicating highly unstable optimization.
    Bottom: AIME accuracy (average of AIME24 and AIME25) and validation response length in tokens, showing that hard truncation quickly destroys performance while failing to provide a controlled length–accuracy trade-off.
    }
    \label{fig:hard-trunc-failure}
\end{figure}

Second, the hard truncation baseline fails catastrophically. As shown in \Cref{fig:hard-trunc-failure}, forcing the model to truncate responses, even on mastered samples, leads to gradient explosion, entropy loss collapse, and thus a very unstable training process. It even produces longer responses on average, and we identify a meaningless repeated pattern at the end or middle of many outputs. This provides strong evidence for our method's design principles: compression must be \textit{soft} (penalizing, not truncating).

Third, we observe an accuracy--efficiency trade-off of a single compression phase, as shown in \Cref{tab:compression-tradeoff}.

\begin{table}[t]
    \centering
    \begin{tabular}{lcccc}
        \toprule
        Step & 170 & 260 & 330 & 385 \\
        \midrule
        Avg.~length reduction & $18.8\%$ & $28.4\%$ & $34.2\%$ & $37.5\%$ \\
        Accuracy drop (points) & $0.3$ & $1.5$ & $2.5$ & $2.7$ \\
        \bottomrule
    \end{tabular}
        \caption{
    Accuracy--efficiency trade-off of a single compression phase for our method (\textsc{ours}).
    We report relative reductions in average response length and the corresponding drops in accuracy (in points) at several representative stopping points along training.
    }\label{tab:compression-tradeoff}
\end{table}

\section{Detailed Analysis}

In this section, we conduct a detailed analysis of our proposed method, investigating its training dynamics, generalization capabilities, and failure modes. We perform several key experiments to validate our approach. 
\begin{itemize}
    \item First, we use a MiMo-7B-SFT model~\citep{xiaomiMiMoUnlockingReasoning2025} and apply our compression training using Dr.GPRO~\citep{liuUnderstandingR1ZeroLikeTraining2025}, mixing varying ratios (10\%, 20\%, 40\%, and 90\%) of mastered samples ($\hat{p}(x)=1.0$) and remaining samples with $\hat{p}(x)\in(0.0,1.0)$ during the compression phase.
    \item Second, to assess architectural generalization, we experiment with Qwen3-30B-A3B~\citep{yangQwen3TechnicalReport2025}, a mixture-of-experts (MoE) model. This model is first doing supervised fine-tuning on open-source CoT data to inject the reasoning pattern, then trained in an accuracy-focused phase for 150 steps using Dr.GRPO with a rollout routing replay mechanism~\citep{maStabilizingMoEReinforcement2025}. Then, we undergo two full loops of the accuracy $\rightarrow$ compression curriculum, again with a 40\% compression mixture.
    \item Finally, we train a single accuracy $\rightarrow$ compression loop on the MiMo-7B-RL model using Dr.GPRO with a 40\% mixture during the compression phase.
\end{itemize}

\paragraph{Training loop dynamics}
We first analyze the dynamics of the proposed accuracy $\rightarrow$ compression $\rightarrow$ accuracy training loop, which demonstrates that our curriculum can simultaneously reduce length and improve accuracy, surpassing the performance ceiling of continuous RLVR training. This loop structure proves stable even when repeated. 

\begin{figure}[htb]
    \centering
    \includegraphics[width=.8\textwidth]{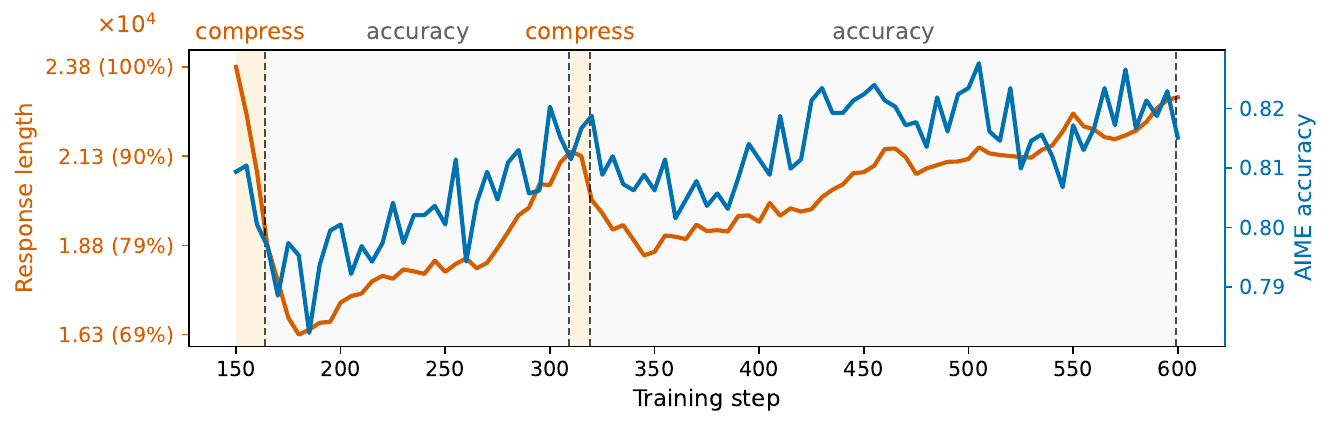}
    \caption{
    Length--accuracy dynamics over two accuracy $\rightarrow$ compression $\rightarrow$ accuracy loops on Qwen3-30B-A3B. The orange curve and left axis show validation response length in tokens (displayed in scientific-notation form with relative percentages with respect to the initial length), while the blue curve and right axis show average AIME accuracy (averaging AIME24 and AIME25). The first loop (150 accuracy steps, 15 compression steps, and 145 accuracy steps) achieves a substantial length reduction with almost no loss in accuracy, and the second loop (10 compression steps and 280 accuracy steps) further improves accuracy while maintaining a net length reduction.
    }
    \label{fig:qwen-loop-length-accuracy}
\end{figure}

\begin{figure}[htb]
    \centering
    \includegraphics[width=.8\textwidth]{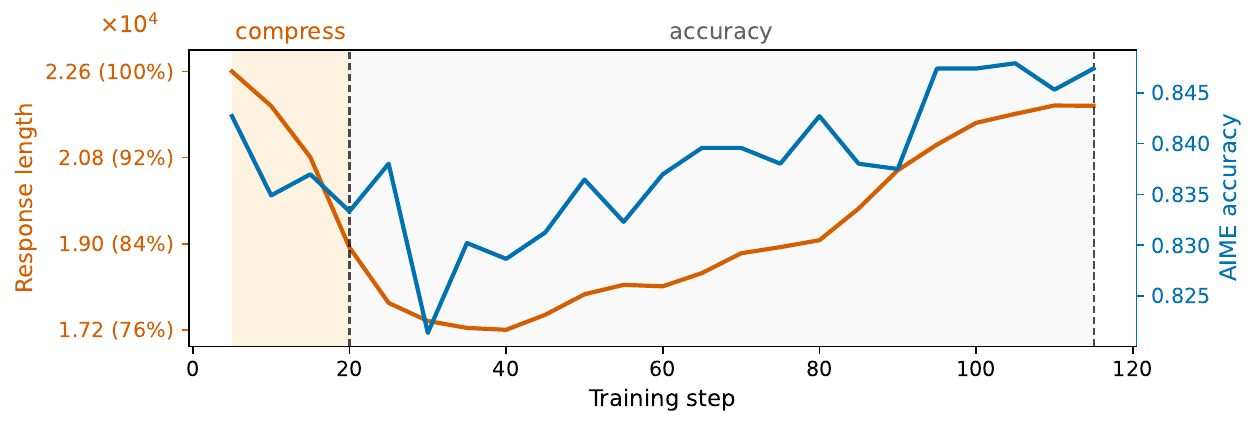}
    \caption{
    Length--accuracy dynamics over a single accuracy $\rightarrow$ compression $\rightarrow$ accuracy loop on MiMo-7B-RL. The orange curve and left axis again show validation response length in tokens (in scientific-notation units and relative percentages), while the blue curve and right axis show average AIME accuracy (averaging AIME24 and AIME25). After roughly 20 compression steps that yield a sizable length reduction with stable accuracy, about 90 subsequent accuracy steps recover and slightly improve accuracy while the length stabilizes at a modest reduction relative to the original.
    }
    \label{fig:mimo-loop-length-accuracy}
\end{figure}

As shown in \Cref{fig:qwen-loop-length-accuracy}, for the Qwen3-30B-A3B model, an initial 150-step accuracy phase was followed by 15 steps of compression, which reduced average response length by 21\% with a negligible accuracy drop (less than 1 point). The subsequent 145-step accuracy phase not only recovered the accuracy but stabilized with a 10\% net length reduction. A second loop (10 compression steps, 280 accuracy steps) achieved the highest accuracy --- a gain of over 1 point from the start --- while maintaining a 3\% final length reduction from the original. Similarly, as shown in \Cref{fig:mimo-loop-length-accuracy}, the MiMo-7B-RL model, after 20 compression steps (16\% length reduction, stable accuracy) and 90 steps of accuracy training, gains about 1 point of accuracy and stabilizes at 97\% of its original length. 

A key observation across models is the effect of optimizer (e.g., from AdamW~\citep{loshchilovDecoupledWeightDecay2018}). After switching from compression back to accuracy, the model briefly continues to shorten responses and may see a transient accuracy drop before the optimization objective stabilizes. For instance, the MiMo-7B model required 15 steps post-compression before accuracy began to recover and 25 steps before the length started to increase.

\paragraph{Cross-domain generalization}
A significant finding is the strong cross-domain generalization of the compression skill. Although compression training was performed \textit{only} on mathematics problems, the model spontaneously reduced response lengths across a diverse range of unseen tasks, often with no loss in performance. In some cases, accuracy improved, which we attribute to mitigating score penalties from excessive-length truncation. 

We evaluated a checkpoint from the main experiment in \Cref{sec:experiments} that had incurred a 3-point accuracy drop on AIME due to compression. Its cross-domain behavior is summarized by the per-benchmark trends in \Cref{fig:benchmark-length-accuracy-grid} and the aggregated statistics in \Cref{tab:benchmark-summary}.

\begin{table}[t]
    \centering
    \begin{tabular}{lccc}
        \toprule
        Benchmark & Category & \makecell[c]{Accuracy\\change (points)} & \makecell[c]{Length\\change (\%)} \\
        \midrule
        \makecell[l]{AIME25\\\citep{MathaiAime25Datasets}} & Math & $-1.7$ & $-31.7$ \\
        \makecell[l]{MATH500\\\citep{HuggingFaceH4MATH500Datasets2025}} & Math & $-0.8$ & $-41.6$ \\
        \makecell[l]{GPQADiamond\\\citep{FingertapGPQADiamondDatasets}} & QA & $-0.4$ & $-34.8$ \\
        \makecell[l]{MMLU\_Redux\\\citep{gemaAreWeDone2025}} & General Knowledge & $+1.1$ & $-33.4$ \\
        \makecell[l]{IFEval\\\citep{zhouInstructionFollowingEvaluationLarge2023}} & Instruction Following & $+0.9$ & $-22.8$ \\
        \makecell[l]{LiveCodeBenchv6\\\citep{jainLiveCodeBenchHolisticContamination2024}} & Code & $+3.7$ & $-31.8$ \\
        \makecell[l]{ZebraLogic\\\citep{linZebraLogicScalingLimits2025}} & Logic & $-1.3$ & $-22.1$ \\
        \bottomrule
    \end{tabular}
    \caption{
    Summary of cross-domain effects of compression at a checkpoint with a 3-point AIME accuracy drop.
    For each benchmark, we report the category, the change in the core metric (in percentage points where the metric is in $[0,1]$, or in raw score points otherwise), and the relative change in average response length (negative values mean shorter responses).
    }
    \label{tab:benchmark-summary}
\end{table}

Two phenomena stand out. On math benchmarks, MATH500 is much easier than AIME25, and we see a stronger length reduction there ($-41.6\%$ vs.\ $-31.7\%$) at similar accuracy, showing that the sample-level policy naturally compresses easier data more aggressively. On LiveCodeBenchv6, accuracy still rises by about 3.7 points despite a $\sim$32\% length reduction because about 18\% of previously overlong, truncated code generations now fall within the budget and receive full credit.

\paragraph{Robustness of compression across algorithms and architectures}
As shown in various experiments throughout this section, we confirmed the robustness of the compression method's behavior. The observed dynamics --- length compression followed by performance recovery --- remained consistent across different GRPO-family algorithms (including Dr.GPRO, DAPO, and rollout routing replay+Dr.GPRO) and diverse model architectures and diverse model sizes, including both the MiMo-7B dense model and the Qwen3-30B-A3B MoE model. 

\begin{figure}[htb]
    \centering
    \begin{subfigure}[t]{.48\textwidth}
        \centering
        \includegraphics[width=\linewidth]{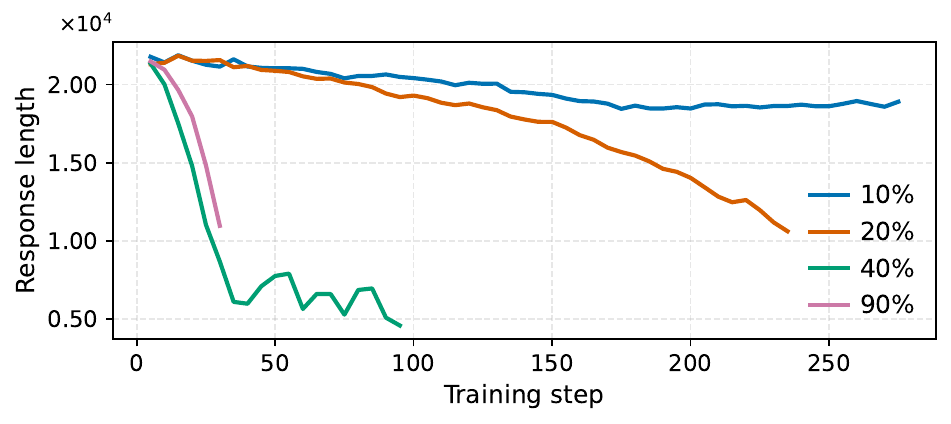}
    \end{subfigure}
    \hfill
    \begin{subfigure}[t]{.48\textwidth}
        \centering
        \includegraphics[width=\linewidth]{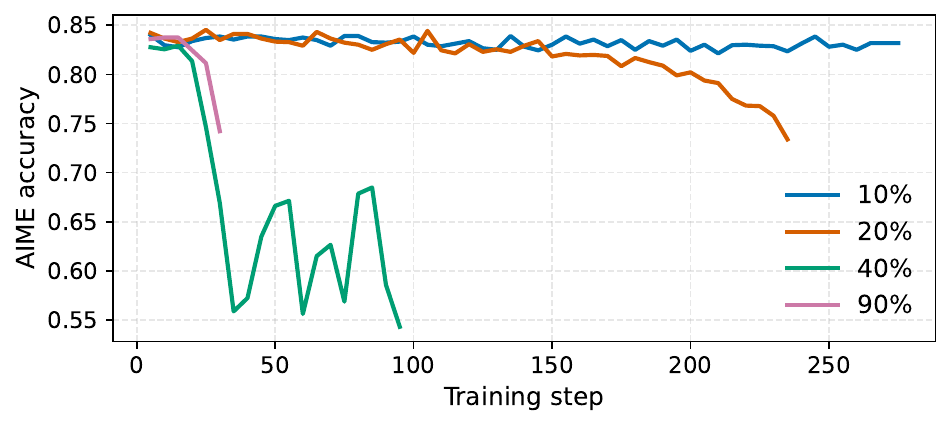}
    \end{subfigure}
    \caption{
    Over-compression dynamics of MiMo-7B-SFT under different compression mixture ratios (10\%, 20\%, 40\%, 90\%).
    Left: validation response length in tokens (scientific-notation y-axis) steadily decreases as compression training continues, with higher mixtures compressing more aggressively.
    Right: AIME accuracy initially remains stable or improves slightly but eventually collapses for all mixtures if compression training is run for too long, with larger mixtures failing earlier, illustrating the universal yet mixture-dependent nature of the over-compression failure mode.
    }
    \label{fig:sft-overcompression}
\end{figure}

\paragraph{Failure mode of over-compression}
We also characterized the failure mode of over-compression, which establishes the necessity of early stopping. As shown in \Cref{fig:sft-overcompression}, in all experiments, regardless of model or data mixture, prolonged compression training inevitably leads to a performance collapse. We observed that higher mixture ratios of compression data accelerate this collapse. The collapsed state is not characterized by repetitive or nonsensical output; rather, the model learns to output incorrect final answers immediately, with no preceding reasoning. 

We analyze this phenomenon as a form of \textit{advantage hacking}. GRPO-style algorithms calculate advantage baselines from on-policy rollouts and do not inherently differentiate between $\hat{p}(x)=0$ and $\hat{p}(x)=1.0$ samples when applying the compression reward. The model discovers a simple hack: intentionally failing a problem ($\hat{p}(x)=0$) while minimizing length yields a better reward than correctly solving it ($\hat{p}(x)=1.0$) and incurring a length penalty. Because the compression skill generalizes so strongly, this hacking behavior, once learned, rapidly propagates. The standard GRPO mechanism exacerbates this, as $\hat{p}(x)=0$ samples contribute no positive gradient signal, offering no penalty for this failure mode.

\paragraph{Case study}
Finally, we conduct a case study of the model's outputs reveals the qualitative impact of this training loop. 

\begin{figure}[htb]
    \centering
    \begin{subfigure}[t]{.48\textwidth}
        \centering
        \includegraphics[width=\linewidth]{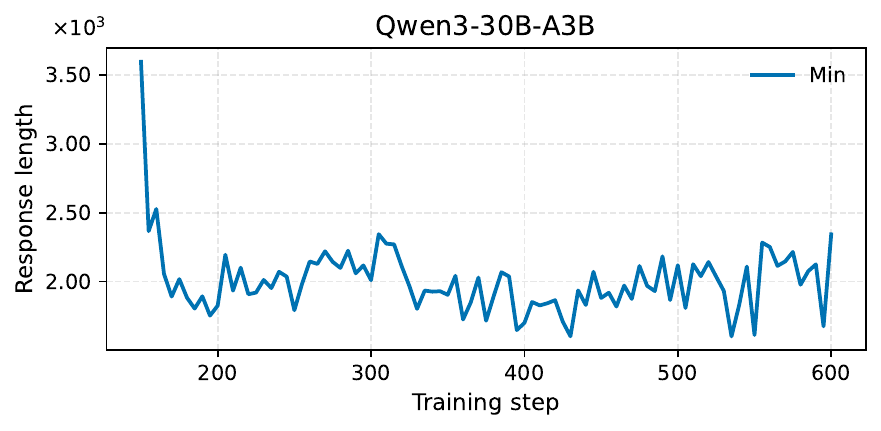}
    \end{subfigure}
    \hfill
    \begin{subfigure}[t]{.48\textwidth}
        \centering
        \includegraphics[width=\linewidth]{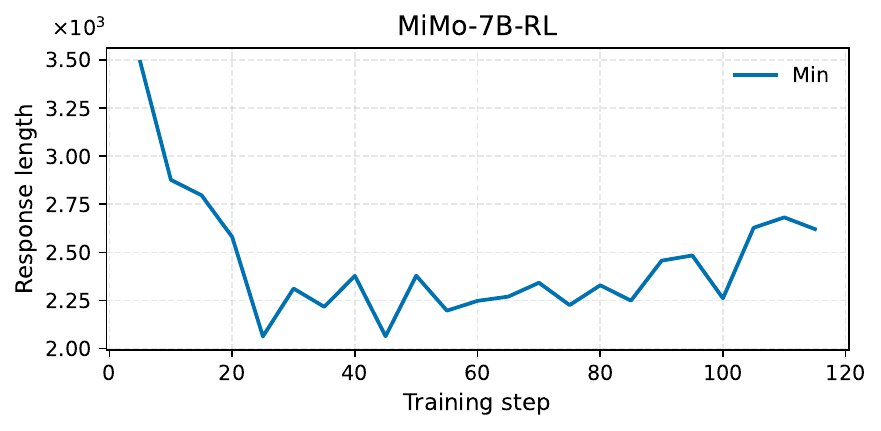}
    \end{subfigure}
    \caption{
    Minimum validation response length over the accuracy $\rightarrow$ compression $\rightarrow$ accuracy loops for Qwen3-30B-A3B (left) and MiMo-7B-RL (right), measured in tokens and plotted with a scientific-notation y-axis.
    In both models, the minimum lengths on easy problems drop sharply during compression and stay low afterwards, while the maximum lengths on difficult problems (not shown) remain near the truncation limit, indicating that the model selectively compresses only where long chains of thought are unnecessary.
    }
    \label{fig:loop-length-span}
\end{figure}

First, the model learns to differentiate between necessary and unnecessary verbosity. As shown in \Cref{fig:loop-length-span}, across various checkpoints during the training loop, the shortest response lengths (on problems it solves easily) drop quickly during compression and \textit{remain} low even after the subsequent accuracy phase, suggesting the model internalizes a new, more efficient ``necessary path.'' Conversely, the longest response lengths (often on unsolved problems) remain long and always being truncated, indicating the model still engages in extended reasoning when faced with difficulty. 

Second, the language style also shifts: the compression phase induces a ``telegraphic'' style, where full sentences are compressed (e.g., ``cosine even, sine odd'') and mathematical formulas are fluently mixed with natural language fragments (e.g., ``So $r_2^{13} = \dots$ = same as before $\dots$''). This compressed style is largely reverted back to more human-like language during the subsequent accuracy phase. We observed this pattern in the Qwen3-30B-A3B model across two full loops. 

\begin{figure}[htb]
    \centering
    \begin{subfigure}[t]{.49\textwidth}
        \centering
        \includegraphics[width=\linewidth]{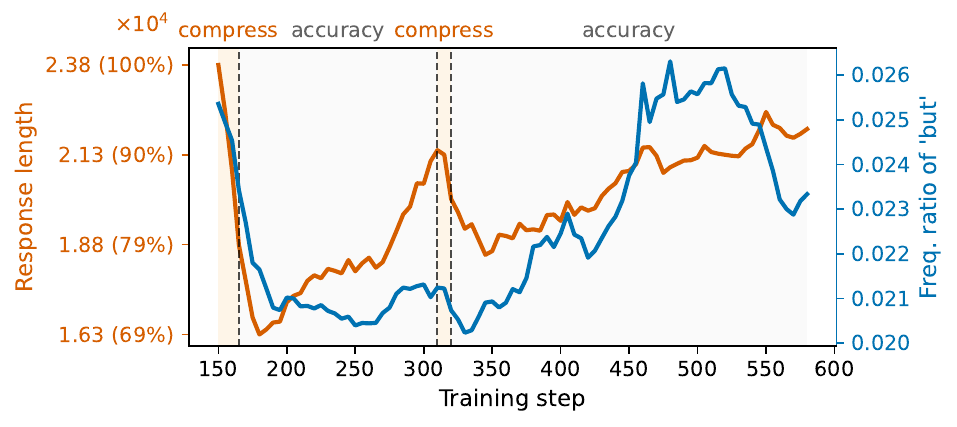}
    \end{subfigure}
    \begin{subfigure}[t]{.49\textwidth}
        \centering
        \includegraphics[width=\linewidth]{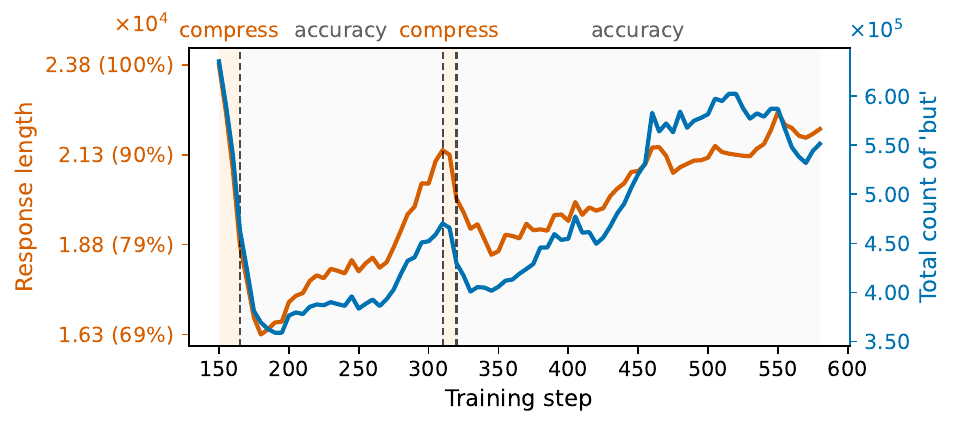}
    \end{subfigure}
    \caption{
    Response length and ``but'' usage over the Qwen3-30B-A3B accuracy $\rightarrow$ compression $\rightarrow$ accuracy loops.
    Left: relative response length in tokens (left axis) and the frequency ratio of ``but'' among generated tokens (right axis).
    Right: the same relative response length and the total count of ``but,'' showing that as compression compresses the chains of thought, transitional markers become less frequent and then rebound during the accuracy recovery phase, closely tracking the length trajectory.
    }
    \label{fig:loop-but-patterns}
\end{figure}

Finally, the reasoning pattern also evolves. As shown in \Cref{fig:loop-but-patterns}, the frequency of transition words (e.g., ``but,'' ``however'') decreases significantly during compression and rises again during accuracy, mirroring the overall length trajectory.

\section{From CoT Compression to Agent Compression}
\label{sec:cot_to_agent}

Our method is motivated by compressing chain-of-thought (CoT) without sacrificing correctness. This framing is not specific to single-shot CoT. In this section, we show that the same recipe directly extends to \textit{agentic} settings, where the model must interleave reasoning with tool use (e.g., edit--test loops), and that compression exhibits strong two-way generalization between non-agent and agent models.

\paragraph{Setup}
We use MiMo-V2-Flash~\citep{teamMiMoV2FlashTechnicalReport2026} as the base model (309B total, 15B active parameters) and run two compression experiments.
\textbf{(i) Non-agent compression:} we train the compression stage on non-agent math, code, and logic tasks for 80 steps.
\textbf{(ii) Agent compression:} starting from a code-agent model optimized for thinking-mode via SFT and RLVR, we train the compression stage on thinking code-agent tasks for 100 steps with tool use enabled.
In both cases, we apply the same mastery-gated RL objective from \Cref{sec:method}, penalizing \textit{response length} (not tool calls or rounds).
We evaluate on three non-agent benchmarks (AIME25, LiveCodeBench-v6, GPQA-Diamond) and one agent benchmark (SWE-Bench Verified~\citep{openaiIntroducingSWEbenchVerified2024}).

\paragraph{Results}
We observe strong two-way generalization between non-agent and agent compression.

\textbf{Non-agent to agent (round reduction without agent training).}
Evaluated on SWE-Bench Verified in \textit{non-thinking} mode, even compression trained purely on non-agent tasks induces more concise agent behavior: the non-thinking SWE agent reduces its average rounds from 110 to 96 ($\Delta R=-13\%$) with only a small accuracy change (73.4 $\rightarrow$ 72.3). This suggests that the learned compression behavior is not confined to single-shot CoT formatting, but can alter the agent's interaction policy.

\textbf{Agent compression (shorter trajectories and cross-domain transfer).}
When we compress the thinking-mode agent directly, we obtain large reductions in both trajectory length and rounds on SWE-Bench Verified: it reduces length from 105k to 34.2k tokens ($\Delta L=-67\%$) and rounds from 96 to 46 ($\Delta R=-52\%$), with a modest accuracy drop (72.1 $\rightarrow$ 70.3). Beyond SWE, the same compressed agent also produces shorter responses on a diverse set of non-agent benchmarks, indicating strong transfer of the learned computation-compression behavior (\Cref{tab:agent_compression_transfer}).

\begin{table}[ht]
\centering
\begin{tabular}{ccc}
\toprule
Benchmark & base & step-100 \\
\midrule
AIME25
& 86.5/29.3k & 85.3/21.9k \,(\,$\Delta L=-25\%$\,) \\
LiveCodeBench-AA
& 76.2/23.3k & 76.0/17.7k \,(\,$\Delta L=-24\%$\,) \\
GPQA-Diamond
& 83.3/24.9k & 84.5/14.0k \,(\,$\Delta L=-44\%$\,) \\
SWE-Bench Verified
& 72.1/105.0k/96 & 70.3/34.2k/46 \,(\,$\Delta L=-67\%$, $\Delta R=-52\%$\,) \\
\bottomrule
\end{tabular}
\vspace{2pt}
\caption{Agent compression: \textbf{accuracy/response-length} on non-agent benchmarks and \textbf{accuracy/response-length/rounds} on SWE-Bench Verified in thinking mode. Relative changes ($\Delta L$, $\Delta R$) are computed against the base checkpoint.}
\label{tab:agent_compression_transfer}
\end{table}

We return, at the very last, to our opening claim that intelligence manifests as compression. Borges writes in \textit{Funes the Memorious}, ``\textit{To think is to forget differences, to generalize, to abstract.}'' In our mastery-gated objective, the model learns first to spend computation where it is needed, and then to \textit{forget} what does not matter: not only to compress chains of thought on mastered problems, but to compress the \textit{entire} reasoning process---tokens, turns, and tool-use trajectories---into a sharper policy of action. On this view, compression is not a cosmetic shortening of explanations; it is a learned allocation of attention and effort. And generalization is the same act, lifted across tasks: preserving the invariants, discarding the incidental, and forgetting the right differences well.

\bibliography{ref}

@inproceedings{aggarwalL1ControllingHow2025,
  title = {L1: {{Controlling How Long A Reasoning Model Thinks With Reinforcement Learning}}},
  shorttitle = {L1},
  booktitle = {Second {{Conference}} on {{Language Modeling}}},
  year = 2025,
  month = aug,
  url = {https://openreview.net/forum?id=4jdIxXBNve#discussion},
  urldate = {2025-11-06},
  author = {Aggarwal, Pranjal and Welleck, Sean}
}

@inproceedings{daiCaptureKeyReasoning2025,
  title = {Capture the {{Key}} in {{Reasoning}} to {{Enhance CoT Distillation Generalization}}},
  booktitle = {Proceedings of the 63rd {{Annual Meeting}} of the {{Association}} for {{Computational Linguistics}} ({{Volume}} 1: {{Long Papers}})},
  year = 2025,
  month = jul,
  pages = {441--465},
  publisher = {Association for Computational Linguistics},
  address = {Vienna, Austria},
  doi = {10.18653/v1/2025.acl-long.21},
  url = {https://aclanthology.org/2025.acl-long.21/},
  urldate = {2025-11-06},
  isbn = {979-8-89176-251-0},
  author = {Dai, Chengwei and Li, Kun and Zhou, Wei and Hu, Songlin},
  editor = {Che, Wanxiang and Nabende, Joyce and Shutova, Ekaterina and Pilehvar, Mohammad Taher}
}

@misc{FingertapGPQADiamondDatasets,
  title = {Fingertap/{{GPQA-Diamond}} · {{Datasets}} at {{Hugging Face}}},
  url = {https://huggingface.co/datasets/fingertap/GPQA-Diamond},
  urldate = {2025-11-14},
  note = {(accessed 2025-11-14)},
  howpublished = {\url{https://huggingface.co/datasets/fingertap/GPQA-Diamond}}
}

@inproceedings{gemaAreWeDone2025,
  title = {Are {{We Done}} with {{MMLU}}?},
  booktitle = {Proceedings of the 2025 {{Conference}} of the {{Nations}} of the {{Americas Chapter}} of the {{Association}} for {{Computational Linguistics}}: {{Human Language Technologies}} ({{Volume}} 1: {{Long Papers}})},
  year = 2025,
  month = apr,
  pages = {5069--5096},
  publisher = {Association for Computational Linguistics},
  address = {Albuquerque, New Mexico},
  doi = {10.18653/v1/2025.naacl-long.262},
  url = {https://aclanthology.org/2025.naacl-long.262/},
  urldate = {2025-11-14},
  isbn = {979-8-89176-189-6},
  author = {Gema, Aryo Pradipta and Leang, Joshua Ong Jun and Hong, Giwon and Devoto, Alessio and Mancino, Alberto Carlo Maria and Saxena, Rohit and He, Xuanli and Zhao, Yu and Du, Xiaotang and Ghasemi Madani, Mohammad Reza and Barale, Claire and McHardy, Robert and Harris, Joshua and Kaddour, Jean and Van Krieken, Emile and Minervini, Pasquale},
  editor = {Chiruzzo, Luis and Ritter, Alan and Wang, Lu}
}

@article{guoDeepSeekR1IncentivizesReasoning2025,
  title = {{{DeepSeek-R1}} Incentivizes Reasoning in {{LLMs}} through Reinforcement Learning},
  year = 2025,
  month = sep,
  journal = {Nature},
  volume = {645},
  number = {8081},
  pages = {633--638},
  publisher = {Nature Publishing Group},
  issn = {1476-4687},
  doi = {10.1038/s41586-025-09422-z},
  url = {https://www.nature.com/articles/s41586-025-09422-z},
  urldate = {2025-11-07},
  copyright = {2025 The Author(s)},
  author = {Guo, Daya and Yang, Dejian and Zhang, Haowei and Song, Junxiao and Wang, Peiyi and Zhu, Qihao and Xu, Runxin and Zhang, Ruoyu and Ma, Shirong and Bi, Xiao and Zhang, Xiaokang and Yu, Xingkai and Wu, Yu and Wu, Z. F. and Gou, Zhibin and Shao, Zhihong and Li, Zhuoshu and Gao, Ziyi and Liu, Aixin and others}
}

@misc{houThinkPrunePruningLong2025,
  title = {{{ThinkPrune}}: {{Pruning Long Chain-of-Thought}} of {{LLMs}} via {{Reinforcement Learning}}},
  shorttitle = {{{ThinkPrune}}},
  year = 2025,
  month = apr,
  number = {arXiv:2504.01296},
  eprint = {2504.01296},
  publisher = {arXiv},
  doi = {10.48550/arXiv.2504.01296},
  url = {http://arxiv.org/abs/2504.01296},
  urldate = {2025-11-06},
  archiveprefix = {arXiv},
  author = {Hou, Bairu and Zhang, Yang and Ji, Jiabao and Liu, Yujian and Qian, Kaizhi and Andreas, Jacob and Chang, Shiyu}
}

@inproceedings{hsiehDistillingStepbyStepOutperforming2023,
  title = {Distilling {{Step-by-Step}}! {{Outperforming Larger Language Models}} with {{Less Training Data}} and {{Smaller Model Sizes}}},
  booktitle = {Findings of the {{Association}} for {{Computational Linguistics}}: {{ACL}} 2023},
  year = 2023,
  month = jul,
  pages = {8003--8017},
  publisher = {Association for Computational Linguistics},
  address = {Toronto, Canada},
  doi = {10.18653/v1/2023.findings-acl.507},
  url = {https://aclanthology.org/2023.findings-acl.507/},
  urldate = {2025-11-06},
  author = {Hsieh, Cheng-Yu and Li, Chun-Liang and Yeh, Chih-kuan and Nakhost, Hootan and Fujii, Yasuhisa and Ratner, Alex and Krishna, Ranjay and Lee, Chen-Yu and Pfister, Tomas},
  editor = {Rogers, Anna and {Boyd-Graber}, Jordan and Okazaki, Naoaki}
}

@misc{HuggingFaceH4MATH500Datasets2025,
  title = {{{HuggingFaceH4}}/{{MATH-500}} · {{Datasets}} at {{Hugging Face}}},
  year = 2025,
  month = nov,
  url = {https://huggingface.co/datasets/HuggingFaceH4/MATH-500},
  urldate = {2025-11-14},
  note = {(accessed 2025-11-14)},
  howpublished = {\url{https://huggingface.co/datasets/HuggingFaceH4/MATH-500}}
}

@inproceedings{jainLiveCodeBenchHolisticContamination2024,
  title = {{{LiveCodeBench}}: {{Holistic}} and {{Contamination Free Evaluation}} of {{Large Language Models}} for {{Code}}},
  shorttitle = {{{LiveCodeBench}}},
  booktitle = {The {{Thirteenth International Conference}} on {{Learning Representations}}},
  year = 2024,
  month = oct,
  url = {https://openreview.net/forum?id=chfJJYC3iL},
  urldate = {2025-11-14},
  author = {Jain, Naman and Han, King and Gu, Alex and Li, Wen-Ding and Yan, Fanjia and Zhang, Tianjun and Wang, Sida and {Solar-Lezama}, Armando and Sen, Koushik and Stoica, Ion}
}

@misc{lanThinkingSpectrumEmpirical2025,
  title = {The {{Thinking Spectrum}}: {{An Empirical Study}} of {{Tunable Reasoning}} in {{LLMs}} through {{Model Merging}}},
  shorttitle = {The {{Thinking Spectrum}}},
  year = 2025,
  month = sep,
  number = {arXiv:2509.22034},
  eprint = {2509.22034},
  publisher = {arXiv},
  doi = {10.48550/arXiv.2509.22034},
  url = {http://arxiv.org/abs/2509.22034},
  urldate = {2025-11-06},
  archiveprefix = {arXiv},
  author = {Lan, Xiaochong and Zheng, Yu and Cao, Shiteng and Li, Yong}
}

@inproceedings{linZebraLogicScalingLimits2025,
  title = {{{ZebraLogic}}: {{On}} the {{Scaling Limits}} of {{LLMs}} for {{Logical Reasoning}}},
  shorttitle = {{{ZebraLogic}}},
  booktitle = {Forty-Second {{International Conference}} on {{Machine Learning}}},
  year = 2025,
  month = jun,
  url = {https://openreview.net/forum?id=sTAJ9QyA6l},
  urldate = {2025-11-14},
  author = {Lin, Bill Yuchen and Bras, Ronan Le and Richardson, Kyle and Sabharwal, Ashish and Poovendran, Radha and Clark, Peter and Choi, Yejin}
}

@inproceedings{liTeachingSmallLanguage2024,
  title = {Teaching {{Small Language Models}} to {{Reason}} for {{Knowledge-Intensive Multi-Hop Question Answering}}},
  booktitle = {Findings of the {{Association}} for {{Computational Linguistics}}: {{ACL}} 2024},
  year = 2024,
  month = aug,
  pages = {7804--7816},
  publisher = {Association for Computational Linguistics},
  address = {Bangkok, Thailand},
  doi = {10.18653/v1/2024.findings-acl.464},
  url = {https://aclanthology.org/2024.findings-acl.464/},
  urldate = {2025-11-06},
  author = {Li, Xiang and He, Shizhu and Lei, Fangyu and JunYang, JunYang and Su, Tianhuang and Liu, Kang and Zhao, Jun},
  editor = {Ku, Lun-Wei and Martins, Andre and Srikumar, Vivek}
}

@misc{liuDLERDoingLength2025,
  title = {{{DLER}}: {{Doing Length pEnalty Right}} - {{Incentivizing More Intelligence}} per {{Token}} via {{Reinforcement Learning}}},
  shorttitle = {{{DLER}}},
  year = 2025,
  month = oct,
  number = {arXiv:2510.15110},
  eprint = {2510.15110},
  publisher = {arXiv},
  doi = {10.48550/arXiv.2510.15110},
  url = {http://arxiv.org/abs/2510.15110},
  urldate = {2025-11-06},
  archiveprefix = {arXiv},
  author = {Liu, Shih-Yang and Dong, Xin and Lu, Ximing and Diao, Shizhe and Liu, Mingjie and Chen, Min-Hung and Yin, Hongxu and Wang, Yu-Chiang Frank and Cheng, Kwang-Ting and Choi, Yejin and Kautz, Jan and Molchanov, Pavlo}
}

@misc{liuUnderstandingR1ZeroLikeTraining2025,
  title = {Understanding {{R1-Zero-Like Training}}: {{A Critical Perspective}}},
  shorttitle = {Understanding {{R1-Zero-Like Training}}},
  year = 2025,
  month = oct,
  number = {arXiv:2503.20783},
  eprint = {2503.20783},
  publisher = {arXiv},
  doi = {10.48550/arXiv.2503.20783},
  url = {http://arxiv.org/abs/2503.20783},
  urldate = {2025-11-07},
  archiveprefix = {arXiv},
  author = {Liu, Zichen and Chen, Changyu and Li, Wenjun and Qi, Penghui and Pang, Tianyu and Du, Chao and Lee, Wee Sun and Lin, Min}
}

@inproceedings{loshchilovDecoupledWeightDecay2018,
  title = {Decoupled {{Weight Decay Regularization}}},
  booktitle = {International {{Conference}} on {{Learning Representations}}},
  year = 2018,
  month = sep,
  url = {https://openreview.net/forum?id=Bkg6RiCqY7},
  urldate = {2025-11-13},
  author = {Loshchilov, Ilya and Hutter, Frank}
}

@misc{luoDeconstructingLongChainofThought2025,
  title = {Deconstructing {{Long Chain-of-Thought}}: {{A Structured Reasoning Optimization Framework}} for {{Long CoT Distillation}}},
  shorttitle = {Deconstructing {{Long Chain-of-Thought}}},
  year = 2025,
  month = mar,
  number = {arXiv:2503.16385},
  eprint = {2503.16385},
  publisher = {arXiv},
  doi = {10.48550/arXiv.2503.16385},
  url = {http://arxiv.org/abs/2503.16385},
  urldate = {2025-11-06},
  archiveprefix = {arXiv},
  author = {Luo, Yijia and Song, Yulin and Zhang, Xingyao and Liu, Jiaheng and Wang, Weixun and Chen, GengRu and Su, Wenbo and Zheng, Bo}
}

@misc{luoO1PrunerLengthHarmonizingFineTuning2025,
  title = {O1-{{Pruner}}: {{Length-Harmonizing Fine-Tuning}} for {{O1-Like Reasoning Pruning}}},
  shorttitle = {O1-{{Pruner}}},
  year = 2025,
  month = jan,
  number = {arXiv:2501.12570},
  eprint = {2501.12570},
  publisher = {arXiv},
  doi = {10.48550/arXiv.2501.12570},
  url = {http://arxiv.org/abs/2501.12570},
  urldate = {2025-11-06},
  archiveprefix = {arXiv},
  author = {Luo, Haotian and Shen, Li and He, Haiying and Wang, Yibo and Liu, Shiwei and Li, Wei and Tan, Naiqiang and Cao, Xiaochun and Tao, Dacheng}
}

@misc{maStabilizingMoEReinforcement2025,
  title = {Stabilizing {{MoE Reinforcement Learning}} by {{Aligning Training}} and {{Inference Routers}}},
  year = 2025,
  month = oct,
  number = {arXiv:2510.11370},
  eprint = {2510.11370},
  publisher = {arXiv},
  doi = {10.48550/arXiv.2510.11370},
  url = {http://arxiv.org/abs/2510.11370},
  urldate = {2025-11-07},
  archiveprefix = {arXiv},
  author = {Ma, Wenhan and Zhang, Hailin and Zhao, Liang and Song, Yifan and Wang, Yudong and Sui, Zhifang and Luo, Fuli}
}

@misc{MathaiAime24Datasets,
  title = {Math-Ai/Aime24 · {{Datasets}} at {{Hugging Face}}},
  url = {https://huggingface.co/datasets/math-ai/aime24},
  urldate = {2025-11-12},
  note = {(accessed 2025-11-12)},
  howpublished = {\url{https://huggingface.co/datasets/math-ai/aime24}}
}

@misc{MathaiAime25Datasets,
  title = {Math-Ai/Aime25 · {{Datasets}} at {{Hugging Face}}},
  url = {https://huggingface.co/datasets/math-ai/aime25},
  urldate = {2025-11-12},
  note = {(accessed 2025-11-12)},
  howpublished = {\url{https://huggingface.co/datasets/math-ai/aime25}}
}

@misc{openaiIntroducingSWEbenchVerified2024,
  title = {Introducing {{SWE-bench Verified}}},
  year = 2024,
  month = aug,
  url = {https://openai.com/index/introducing-swe-bench-verified/},
  urldate = {2026-01-19},
  note = {(accessed 2026-01-19)},
  howpublished = {\url{https://openai.com/index/introducing-swe-bench-verified/}},
  author = {{OpenAI}}
}

@misc{shaoDeepSeekMathPushingLimits2024,
  title = {{{DeepSeekMath}}: {{Pushing}} the {{Limits}} of {{Mathematical Reasoning}} in {{Open Language Models}}},
  shorttitle = {{{DeepSeekMath}}},
  year = 2024,
  month = apr,
  number = {arXiv:2402.03300},
  eprint = {2402.03300},
  publisher = {arXiv},
  doi = {10.48550/arXiv.2402.03300},
  url = {http://arxiv.org/abs/2402.03300},
  urldate = {2025-11-07},
  archiveprefix = {arXiv},
  author = {Shao, Zhihong and Wang, Peiyi and Zhu, Qihao and Xu, Runxin and Song, Junxiao and Bi, Xiao and Zhang, Haowei and Zhang, Mingchuan and Li, Y. K. and Wu, Y. and Guo, Daya}
}

@misc{teamKimiK15Scaling2025,
  title = {Kimi K1.5: {{Scaling Reinforcement Learning}} with {{LLMs}}},
  shorttitle = {Kimi K1.5},
  year = 2025,
  month = jun,
  number = {arXiv:2501.12599},
  eprint = {2501.12599},
  publisher = {arXiv},
  doi = {10.48550/arXiv.2501.12599},
  url = {http://arxiv.org/abs/2501.12599},
  urldate = {2025-11-06},
  archiveprefix = {arXiv},
  author = {Team, Kimi and Du, Angang and Gao, Bofei and Xing, Bowei and Jiang, Changjiu and Chen, Cheng and Li, Cheng and Xiao, Chenjun and Du, Chenzhuang and Liao, Chonghua and Tang, Chuning and Wang, Congcong and Zhang, Dehao and Yuan, Enming and Lu, Enzhe and Tang, Fengxiang and Sung, Flood and Wei, Guangda and Lai, Guokun and others}
}

@misc{teamMiMoV2FlashTechnicalReport2026,
  title = {{{MiMo-V2-Flash Technical Report}}},
  year = 2026,
  month = jan,
  number = {arXiv:2601.02780},
  eprint = {2601.02780},
  publisher = {arXiv},
  doi = {10.48550/arXiv.2601.02780},
  url = {http://arxiv.org/abs/2601.02780},
  urldate = {2026-01-19},
  archiveprefix = {arXiv},
  author = {Team, Xiaomi LLM-Core and Xiao, Bangjun and Xia, Bingquan and Yang, Bo and Gao, Bofei and Shen, Bowen and Zhang, Chen and He, Chenhong and Lou, Chiheng and Luo, Fuli and Wang, Gang and Xie, Gang and Zhang, Hailin and Lv, Hanglong and Li, Hanyu and Chen, Heyu and Xu, Hongshen and Zhang, Houbin and Liu, Huaqiu and others}
}

@misc{wangEfficientLongCoT2025,
  title = {Efficient {{Long CoT Reasoning}} in {{Small Language Models}}},
  year = 2025,
  month = jun,
  number = {arXiv:2505.18440},
  eprint = {2505.18440},
  publisher = {arXiv},
  doi = {10.48550/arXiv.2505.18440},
  url = {http://arxiv.org/abs/2505.18440},
  urldate = {2025-11-06},
  archiveprefix = {arXiv},
  author = {Wang, Zhaoyang and Jiang, Jinqi and Qiu, Tian and Liu, Hui and Tang, Xianfeng and Yao, Huaxiu}
}

@inproceedings{weiChainofthoughtPromptingElicits2022,
  title = {Chain-of-Thought Prompting Elicits Reasoning in Large Language Models},
  booktitle = {Proceedings of the 36th {{International Conference}} on {{Neural Information Processing Systems}}},
  year = 2022,
  month = nov,
  series = {{{NIPS}} '22},
  pages = {24824--24837},
  publisher = {Curran Associates Inc.},
  address = {Red Hook, NY, USA},
  urldate = {2025-11-06},
  isbn = {978-1-7138-7108-8},
  author = {Wei, Jason and Wang, Xuezhi and Schuurmans, Dale and Bosma, Maarten and Ichter, Brian and Xia, Fei and Chi, Ed H. and Le, Quoc V. and Zhou, Denny}
}

@misc{wuUnlockingEfficientLongtoShort2025,
  title = {Unlocking {{Efficient Long-to-Short LLM Reasoning}} with {{Model Merging}}},
  year = 2025,
  month = may,
  number = {arXiv:2503.20641},
  eprint = {2503.20641},
  publisher = {arXiv},
  doi = {10.48550/arXiv.2503.20641},
  url = {http://arxiv.org/abs/2503.20641},
  urldate = {2025-11-06},
  archiveprefix = {arXiv},
  author = {Wu, Han and Yao, Yuxuan and Liu, Shuqi and Liu, Zehua and Fu, Xiaojin and Han, Xiongwei and Li, Xing and Zhen, Hui-Ling and Zhong, Tao and Yuan, Mingxuan}
}

@misc{xiaomiMiMoUnlockingReasoning2025,
  title = {{{MiMo}}: {{Unlocking}} the {{Reasoning Potential}} of {{Language Model}} -- {{From Pretraining}} to {{Posttraining}}},
  shorttitle = {{{MiMo}}},
  year = 2025,
  month = jun,
  number = {arXiv:2505.07608},
  eprint = {2505.07608},
  publisher = {arXiv},
  doi = {10.48550/arXiv.2505.07608},
  url = {http://arxiv.org/abs/2505.07608},
  urldate = {2025-11-12},
  archiveprefix = {arXiv},
  author = {Xiaomi, LLM-Core and Xia, Bingquan and Shen, Bowen and Cici and Zhu, Dawei and Zhang, Di and Wang, Gang and Zhang, Hailin and Liu, Huaqiu and Xiao, Jiebao and Dong, Jinhao and Zhao, Liang and Li, Peidian and Wang, Peng and Yu, Shihua and Chen, Shimao and Wang, Weikun and Ma, Wenhan and Deng, Xiangwei and others}
}

@inproceedings{xiaTokenSkipControllableChainofThought2025,
  title = {{{TokenSkip}}: {{Controllable Chain-of-Thought Compression}} in {{LLMs}}},
  shorttitle = {{{TokenSkip}}},
  booktitle = {Proceedings of the 2025 {{Conference}} on {{Empirical Methods}} in {{Natural Language Processing}}},
  year = 2025,
  month = nov,
  pages = {3351--3363},
  publisher = {Association for Computational Linguistics},
  address = {Suzhou, China},
  url = {https://aclanthology.org/2025.emnlp-main.165/},
  urldate = {2025-11-06},
  isbn = {979-8-89176-332-6},
  author = {Xia, Heming and Leong, Chak Tou and Wang, Wenjie and Li, Yongqi and Li, Wenjie},
  editor = {Christodoulopoulos, Christos and Chakraborty, Tanmoy and Rose, Carolyn and Peng, Violet}
}

@misc{yangQwen3TechnicalReport2025,
  title = {Qwen3 {{Technical Report}}},
  year = 2025,
  month = may,
  number = {arXiv:2505.09388},
  eprint = {2505.09388},
  publisher = {arXiv},
  doi = {10.48550/arXiv.2505.09388},
  url = {http://arxiv.org/abs/2505.09388},
  urldate = {2025-11-13},
  archiveprefix = {arXiv},
  author = {Yang, An and Li, Anfeng and Yang, Baosong and Zhang, Beichen and Hui, Binyuan and Zheng, Bo and Yu, Bowen and Gao, Chang and Huang, Chengen and Lv, Chenxu and Zheng, Chujie and Liu, Dayiheng and Zhou, Fan and Huang, Fei and Hu, Feng and Ge, Hao and Wei, Haoran and Lin, Huan and Tang, Jialong and others}
}

@inproceedings{yuDAPOOpenSourceLLM2025,
  title = {{{DAPO}}: {{An Open-Source LLM Reinforcement Learning System}} at {{Scale}}},
  shorttitle = {{{DAPO}}},
  booktitle = {The {{Thirty-ninth Annual Conference}} on {{Neural Information Processing Systems}}},
  year = 2025,
  month = oct,
  url = {https://openreview.net/forum?id=2a36EMSSTp&referrer=%5Bthe%20profile%20of%20Ruofei%20Zhu%5D(%2Fprofile%3Fid%3D~Ruofei_Zhu1)},
  urldate = {2025-11-06},
  author = {Yu, Qiying and Zhang, Zheng and Zhu, Ruofei and Yuan, Yufeng and Zuo, Xiaochen and YuYue and Dai, Weinan and Fan, Tiantian and Liu, Gaohong and Liu, Juncai and Liu, LingJun and Liu, Xin and Lin, Haibin and Lin, Zhiqi and Ma, Bole and Sheng, Guangming and Tong, Yuxuan and Zhang, Chi and Zhang, Mofan and others}
}

@misc{zhangLightThinkerThinkingStepbyStep2025,
  title = {{{LightThinker}}: {{Thinking Step-by-Step Compression}}},
  shorttitle = {{{LightThinker}}},
  year = 2025,
  month = sep,
  number = {arXiv:2502.15589},
  eprint = {2502.15589},
  publisher = {arXiv},
  doi = {10.48550/arXiv.2502.15589},
  url = {http://arxiv.org/abs/2502.15589},
  urldate = {2025-11-06},
  archiveprefix = {arXiv},
  author = {Zhang, Jintian and Zhu, Yuqi and Sun, Mengshu and Luo, Yujie and Qiao, Shuofei and Du, Lun and Zheng, Da and Chen, Huajun and Zhang, Ningyu}
}

@misc{zhaoCanPruningImprove2025,
  title = {Can {{Pruning Improve Reasoning}}? {{Revisiting Long-CoT Compression}} with {{Capability}} in {{Mind}} for {{Better Reasoning}}},
  shorttitle = {Can {{Pruning Improve Reasoning}}?},
  year = 2025,
  month = aug,
  number = {arXiv:2505.14582},
  eprint = {2505.14582},
  publisher = {arXiv},
  doi = {10.48550/arXiv.2505.14582},
  url = {http://arxiv.org/abs/2505.14582},
  urldate = {2025-11-06},
  archiveprefix = {arXiv},
  author = {Zhao, Shangziqi and Yuan, Jiahao and Yang, Guisong and Naseem, Usman}
}

@misc{zhengGroupSequencePolicy2025,
  title = {Group {{Sequence Policy Optimization}}},
  year = 2025,
  month = jul,
  number = {arXiv:2507.18071},
  eprint = {2507.18071},
  publisher = {arXiv},
  doi = {10.48550/arXiv.2507.18071},
  url = {http://arxiv.org/abs/2507.18071},
  urldate = {2025-11-07},
  archiveprefix = {arXiv},
  author = {Zheng, Chujie and Liu, Shixuan and Li, Mingze and Chen, Xiong-Hui and Yu, Bowen and Gao, Chang and Dang, Kai and Liu, Yuqiong and Men, Rui and Yang, An and Zhou, Jingren and Lin, Junyang}
}

@misc{zhouInstructionFollowingEvaluationLarge2023,
  title = {Instruction-{{Following Evaluation}} for {{Large Language Models}}},
  year = 2023,
  month = nov,
  number = {arXiv:2311.07911},
  eprint = {2311.07911},
  publisher = {arXiv},
  doi = {10.48550/arXiv.2311.07911},
  url = {http://arxiv.org/abs/2311.07911},
  urldate = {2025-11-14},
  archiveprefix = {arXiv},
  author = {Zhou, Jeffrey and Lu, Tianjian and Mishra, Swaroop and Brahma, Siddhartha and Basu, Sujoy and Luan, Yi and Zhou, Denny and Hou, Le}
}

\end{document}